\documentclass[journal,draftcls,onecolumn,12pt,twoside]{IEEEtranTCOM}
\normalsize

\usepackage{cite}

\usepackage{algorithm}
\usepackage{algpseudocode}

\ifCLASSINFOpdf
  \usepackage[pdftex]{graphicx}
\else
\fi

\usepackage{amsmath, amsthm, amssymb, multirow}

\newcommand{\Rg}{\mathcal{R}}
\newcommand{\Ind}{\mathbf{1}}

\begin{document}
%
\pagestyle{plain}
\title{Impulsive Noise Mitigation \\ in Powerline Communications \\ Using Sparse Bayesian Learning}

%
%

\author{Jing~Lin,~\IEEEmembership{Student Member,~IEEE,}
        Marcel~Nassar,~\IEEEmembership{Student Member,~IEEE,}
        and~Brian~L.~Evans,~\IEEEmembership{Fellow,~IEEE}
\thanks{J. Lin, M. Nassar and B. L. Evans are with the Department
of Electrical and Computer Engineering, The University of Texas at Austin, Austin,
TX, 78712 USA e-mail: linj@mail.utexas.edu, mnassar@utexas.edu, bevans@ece.utexas.edu.}
\thanks{This work was supported by the Global Research Collaboration Program of the Semiconductor Research Corporation under Task Id 1836.063, and was presented in part at the 2011 IEEE International Global Communications Conference \cite{lin2011non}.}}

%
%

\markboth{To appear in IEEE Journal on Selected Areas of Communications}%
{Submitted paper}
%




\maketitle
\begin{abstract}
Additive asynchronous and cyclostationary impulsive noise limits communication performance in OFDM powerline communication (PLC) systems. Conventional OFDM receivers assume additive white Gaussian noise and hence experience degradation in communication performance in impulsive noise.  Alternate designs assume a parametric statistical model of impulsive noise and use the model parameters in mitigating impulsive noise. These receivers require overhead in training and parameter estimation, and degrade due to model and parameter mismatch, especially in highly dynamic environments. In this paper, we model impulsive noise as a sparse vector in the time domain without any other assumptions, and apply sparse Bayesian learning methods for estimation and mitigation without training. 
We propose three iterative algorithms with different complexity vs. performance trade-offs: (1) we utilize the noise projection onto null and pilot tones to estimate and subtract the noise impulses; (2) we add the information in the data tones to perform  joint noise estimation and OFDM detection; (3) we embed our algorithm into a decision feedback structure to further enhance the performance of coded systems.
When compared to conventional OFDM PLC receivers, the proposed receivers achieve SNR gains of up to 9 dB in coded and 10 dB in uncoded systems in the presence of impulsive noise.
\end{abstract}

\begin{IEEEkeywords}
Asynchronous impulsive noise, cyclostationary noise, PLC, OFDM, sparse Bayesian learning.
\end{IEEEkeywords}

%
\IEEEpeerreviewmaketitle

\section{Introduction}
\label{sec:intro}

Unlike traditional electric grids that carry one-way flow of power from generators to customers, smart grids use two-way flows of information to create an intelligent energy delivery network. Various technologies have emerged to facilitate data communications throughout the grid, especially in two scenarios: outdoor communications between local utilities and customers, and indoor communications for home area networks. The purpose is to support smart grid applications such as smart metering and real-time energy management by the utility and the home users.

Powerline communications (PLC) have been attractive as a solution for smart grid communications. For outdoor communications, there has been a lot of interest in developing narrowband (NB) PLC systems that operate in the 3--500 kHz band to provide data rates up to 800 kbps \cite{Nassar2012}. Examples of NB PLC systems are specified in the industry developed standards PRIME \cite{prime} and G3 \cite{g3}, and recent international standards IEEE P1901.2 and ITU-T G.hnem. For indoor communications, broadband (BB) PLC systems that target higher data rates (up to 200 Mbps) utilizing the 1.8--250 MHz band have been standardized in IEEE P1901 and ITU-T G.hn. 

One of the major impairments in PLC systems is the additive non-Gaussian powerline noise. The noise can generally be decomposed into spectrally-shaped background noise, cyclostationary impulsive noise, and asynchronous impulsive noise \cite{zimmermann2000analysis}. The impulsive noise consists of short bursts of noise with power spectral density significantly higher than background noise. The ocurrence of noise bursts can be periodic or random, hence the terms ``cyclostationary" and ``asynchronous", respectively. Recent field measurements on outdoor medium-voltage (MV) and low-voltage (LV) power lines have shown that cyclostationary impulsive noise synchronous to half the AC main's cycle is the dominant noise component in the 3--500 kHz band \cite{nassar2012cyclostationary}. In addition, the ever-increasing deployment of heterogeneous PLC systems is leading to severe co-channel interference that is asynchronous impulsive noise in nature \cite{nassar2011statistical}. The asynchronous impulsive noise is also more significant in indoor BB PLC systems due to the switching transients of electrical appliances \cite{zimmermann2000analysis}.

Orthogonal frequency division multiplexing (OFDM) is adopted in most modern PLC standards, primarily due to its advantages in combating frequency-selective channels and spectrally-shaped noise. The presence of noise bursts as much as 50 dB above background noise  limits the overall channel capacity by decreasing the signal-to-noise ratio (SNR) in certain frequency band and time durations \cite{zimmermann2000analysis,nassar2012cyclostationary}. 
Various statistical properties of the noise can be utilized to improve the reliability and throughput  of the communication system in impulsive noise environments. In particular, these statistics can be used to tune the performance of the transmitter and receiver to the specific noise scenario represented by those statistics \cite{haringThesis,cho2004joint,yoo2008asymptotic}. Although such an approach performs well in the specific noise scenario, it is inflexible and requires significant design effort. A simpler and effective approach is to denoise the communication signal before detection. In this approach, a noise estimation block would provide a noise estimate and subtract it from the received signal before passing it to the conventional OFDM receiver. 

In this work we aim to mitigate both asynchronous and cyclostationary impulsive noise in OFDM PLC receivers by proposing three denoising algorithms based on sparse Bayesian learning (SBL) techniques \cite{wipf2004sparse}. We first target the impulsive noise whose  bursts are much shorter than an OFDM symbol (e.g. asynchronous impulsive noise). This noise is sparse in the time domain and can be estimated using SBL either by observing non-data tones such as nulls and pilots or by jointly decoding across all tones.
Since the algorithms do not make any assumptions on the noise models and do not require extra training, we refer to them as ``non-parametric'' techniques. 
To deal with the impulsive noise with bursts spanning over one or more OFDM symbols (e.g. cyclostationary impulsive noise), we adopt a time-domain interleaving OFDM transceiver structure \cite{dweik} where the transmitted and received signals are interleaved and deinterleaved, respectively, in the time domain across multiple OFDM symbols. The effect of the deinterleaver is to spread the long noise bursts into shorter ones. After deinterleaving, noise within each OFDM symbol becomes sparse in the time domain and we can apply the proposed denoising methods.

The rest of the paper is organized as follows. In Section \ref{sec:ImpNoiseEnv} we discuss the nature of both asynchronous and cyclostationary impulsive noise in PLC networks and the corresponding statistical noise models. Although our proposed methods are noise model independent, these models are useful for simulating various impulsive noise scenarios to test our algorithms. We review existing receiver algorithms for impulsive noise mitigation in Section \ref{sec:RelatedWork}. Having established the system model in Section \ref{sec:SysModel}, we introduce the three non-parametric impulsive noise mitigation algorithms in Section \ref{sec:impulsiveNoiseEst}. Then we perform complexity analysis and present a low-complexity implementation of the first proposed algorithm in Section \ref{sec:SequentialSBLImplementation}. To demonstrate the performance of our algorithms, simulation results are presented and discussed in Section \ref{sec:SimulationResults}.

\section{Statistical Modeling of Impulsive Noise in PLC}
\label{sec:ImpNoiseEnv}
Noise in communication systems is typically modeled as a Gaussian random process. While an appropriate model for the thermal noise in the receiver circuitry, it fails to capture the characteristics of the noise and interference in PLC networks \cite{zimmermann2000analysis}. Various power electronic devices, especially those with switching circuitry, inject random or periodic emissions of noise into the connected power lines.
The resulting noise deviates from the Gaussian model and is impulsive in nature. Depending on the operating frequency range, we distinguish between two types of impulsive noise in PLC: the asynchronous and the cyclostationary impulse noise.

\subsection{Asynchronous Impulsive Noise Modeling}
\label{subsec:asynchronousNoiseModel}

\begin{figure*}[t]
  \centering
  \includegraphics[width=15cm]{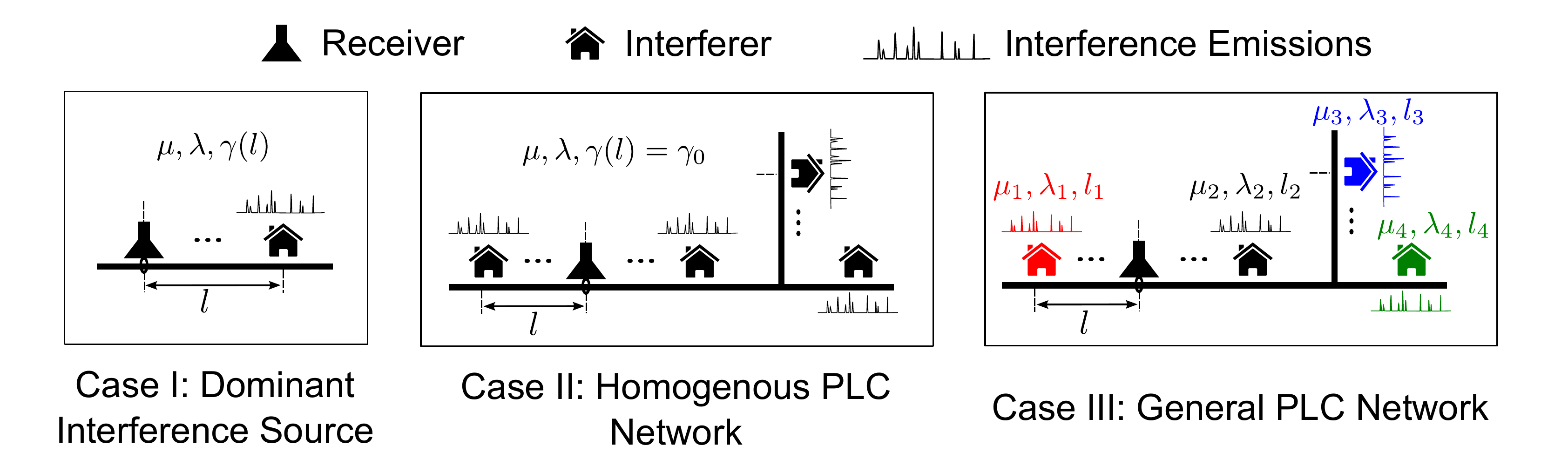}
\caption{Interference scenarios in PLC networks. Each interferer emits a random sequence of emissions onto the powe line, which add up at the receiver. Each interferer is described statistically by a mean number of emission events $\mu$, mean duration between emission events $\lambda$, and the pathloss to the receiver $\gamma$.}
\label{fig:networkModels}
\end{figure*}

In PLC networks, asynchronous impulsive noise arises from random emission events occurring at residential and industrial sites. Typical interference sources include appliance switching transients and uncoordinated PLC transmissions due to co-existence issues. This type of noise is dominant in the higher frequency ranges from several hundred kHz to 20~MHz \cite{Zimmermann2002a}. Various studies have empirically fitted the first order statistics of the noise to the Gaussian mixture, Middleton Class A, or Symmetric Alpha Stable (S$\alpha$S) models \cite{Chan1989}. Statistical-physical models of the asynchronous impulsive noise have been analytically derived based on  PLC network topologies, and by modeling the random emissions as a temporal Poisson point process \cite{nassar2011statistical}. The three scenarios analyzed in \cite{nassar2011statistical} are given in \figurename~\ref{fig:networkModels} and \tablename~\ref{tab:networkModels} along with the first order statistics of the asynchronous impulsive noise.

For convenience of the discussion in later sections, we briefly describe the Gaussian mixture and the Middleton Class A models as follows.
\subsubsection{Gaussian Mixture Model}
A random variable $Z$ has a Gaussian mixture distribution if its probability density function (pdf) is a weighted summation of different Gaussian distributions
\begin{equation}
	f_Z(z)=\sum_{k=0}^{K}{\pi_k\cdot \mathcal{N}(z;0,\gamma_k)},
\label{GMpdf}
\end{equation}
where $\mathcal{N}(z;0,\gamma_k)$ denotes a Gaussian pdf with zero mean and variance $\gamma_k$, and $\pi_k$ is the mixing probability of the $k$-th Gaussian component. 

\subsubsection{Middleton Class A Model}
Middleton Class A model \cite{middleton2007statistical}, with an overlapping factor $A\in[10^{-2},1]$ and power ratio $\Omega\in[10^{-6},1]$, can be considered as a special case of the Gaussian mixture distribution, with $\pi_k=e^{-A}\frac{A^k}{k!}$ and  $\gamma_k=\frac{k/A+\Omega}{1+\Omega}$ as $K\rightarrow \infty$. In practice, only the first few significant terms are retained.




\begin{table}[h]
  \centering
  	\begin{tabular}{ | c | c | c | }
  	\hline
  	\bfseries{Scenario} & \bfseries{Example Network} & \bfseries{Statistical Model} \\ \hline
  	\multirow{2}{*}{Dominant Interferer} & Rural Area & Middleton Class A: \\
	 & Industrial Area & $A=\lambda\mu$ , $\Omega=A\gamma E\left[h^2 B^2 \right]/2$ \\ \hline
  	\multirow{2}{*}{Homogeneous PLC Network} & Urban Area & Middleton Class A: \\
	 & Residential Buildings & $A=M\lambda\mu$ , $\Omega=A\gamma E\left[h^2 B^2 \right]/2M$ \\ \hline
	\multirow{2}{*}{General PLC Network} & Dense Urban Area & Gaussian Mixture: \\
	& Commercial & $\pi_k$ and $\gamma_k$ given in \cite{nassar2011statistical} \\ \hline
  	\end{tabular}
  \caption{Statistical-physical models of interference in PLC networks categorized by network types. Parameters are given in \figurename~\ref{fig:networkModels} and $M$ is the number of interferers.}
  \label{tab:networkModels}
\end{table}

\subsection{Cyclostationary Impulsive Noise Modeling}
\label{subsec:cyclostationaryNoiseModel}
\begin{figure}[t]
  \centering
  \includegraphics[width=9cm]{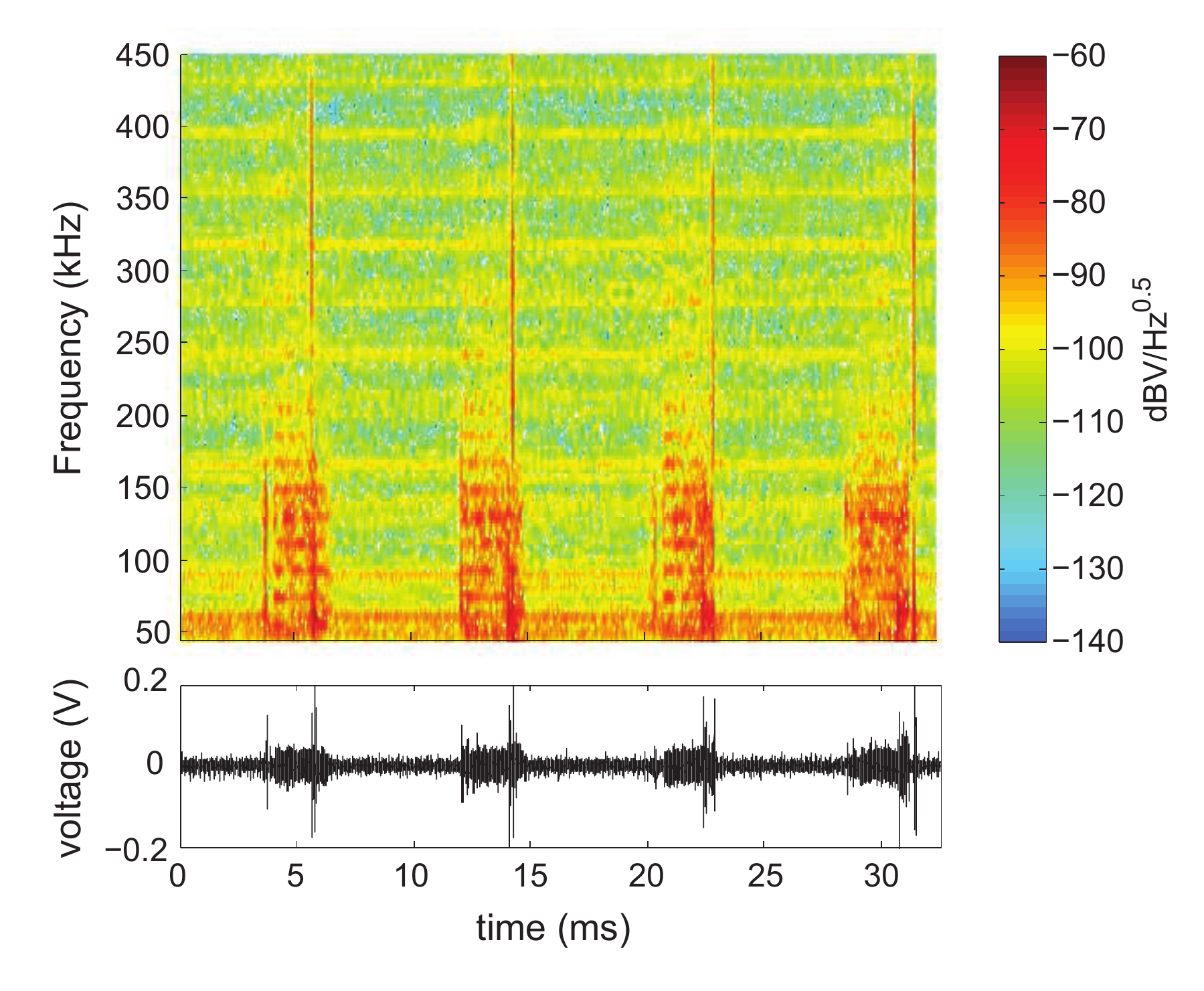}
\caption{The spectrogram of a noise trace at a low-voltage site. It highlights the cyclostationary nature of the noise both in time and frequency domain.}
\label{fig:cycloTrace}
\end{figure}

While asynchronous impulsive noise dominates the higher frequency bands, recent studies showed that cyclostationary noise is the main impairment in the $3-500$~kHz range \cite{Katayama2006,nassar2012cyclostationary,Nassar2012}. The cyclostationary noise has periodically varying statistics with the period equal to half the AC main's cycle.
 A trace collected from a low voltage site is given in \figurename~\ref{fig:cycloTrace}. 

The spectrogram in \figurename~\ref{fig:cycloTrace} indicates that the noise exhibits regions during each period where it appears stationary. This fact was exploited in \cite{nassar2012cyclostationary} to propose a linear periodically time varying (LPTV) system model that divides noise samples in each period into $M$ stationary regions $\{\Rg_i\}_{i=1}^M$. Each region is characterized by a spectral shape and a corresponding linear time-invariant (LTI) spectrum shaping filter.
Given this model, the cyclostationary noise $n[k]$ can be generated as the convolution of an AWGN signal $s[k]$ with an LPTV filter given by
\begin{equation}\label{eq:cyclonoise}
n[k] = \sum\limits_{\tau}h[k,\tau]s[\tau]= \sum\limits_{i=1}^M \Ind_{k \in \Rg_i} \sum\limits_{\tau} h_i[k-\tau] s[\tau]
\end{equation}
where $\Ind_\mathcal{A}$ is the indicator function. 

\section{Prior Work in Impulsive Noise Mitigation}
\label{sec:RelatedWork}

\subsection{Asynchronous Impulsive Noise Mitigation}
\label{subsec:asynchronousNoiseMitigation}

Asynchronous impulsive noise has been widely seen not only in PLC but also in wireless communications systems. In both contexts, various parametric and non-parametric methods have been proposed to mitigate the effect of asynchronous impulsive noise in OFDM systems. 

In parametric methods, the receiver assumes a particular statistical noise model, and typically estimates the parameters of the statistical model during a training stage. For example, pre-filtering techniques based on the Alpha-Stable model have been proposed in \cite{Nassar2009-03-27}. While low in complexity, their performance can deteriorate significantly as the constellation size increases. More promising results were reported in \cite{haringThesis}, where two minimum mean-squared-error (MMSE) symbol-by-symbol estimators were derived assuming Gaussian mixture modeled impulsive noise. Additionally, in \cite{haring2003iterative} an iterative decoding algorithm was introduced to cope with Middleton Class A impulsive noise. The advantage of parametric methods is that they lead to performance gains by exploiting information of the noise model and its parameters. However, they require extra training overhead and can suffer from performance degradation when the noise model and/or the parameters mismatch the possibly time varying statistics of the impulsive noise.

Non-parametric methods, on the other hand, require no additional training or parameter estimation since they do not assume any particular noise model. This makes them more robust to various environments and more resilient to model mismatches. Taking advantage of the sparsity of the time-domain impulsive noise, \cite{caire2008impulse} applied the compressed sensing (CS) techniques to estimate the impulsive noise from the null tones of the received signal. The algorithm was subject to a sufficient recovery condition that the number of impulses within an OFDM symbol does not exceed a threshold that is uniquely determined by the discrete Fourier transform (DFT) size and the number of null tones. However, for common OFDM system settings, the threshold turns out to be too restrictive for many impulsive noise environments where an OFDM symbol is subjected to multiple impulses during its duration. This CS-based approach was extended in \cite{lampe2011bursty} to a bursty impulsive noise detector that exploits the block-sparsity of the noise. The performance of the algorithm, however, is affected by parameters that should be ideally adapted to the number of noise bursts within an OFDM symbol, and the background noise level.

Our work seeks to develop non-parametric asynchronous impulsive noise mitigation algorithms that are applicable to all impulsive noise scenarios. Towards this end, we extend the CS based algorithm in \cite{caire2008impulse} to a sparse Bayesian learning (SBL) approach \cite{wipf2004sparse} for improved performance and robustness.

\subsection{Cyclostationary Impulsive Noise Mitigation}
\label{subsec:cyclostationaryNoiseMitigation}

Early research on cyclostationary signal detection and extraction were primarily based on the theory of cumulant and cyclic spectrum \cite{gardner1994cumulant}. In particular, the cyclic spectrum of a second-order cyclostationary process contains harmonic peaks, which can be used for detection and extraction \cite{bonnardot2005extraction}. Accurate estimation of the cyclic spectrum, however, generally require a large amount of data, which potentially limits its applications to real-time communications systems.

Recent research on transmitter and receiver design has explicitly taken into account of the cyclostationary noise. Some of the work attempted to optimize the transmitted signal and the receiver equalization under cyclostationary noise \cite{cho2004joint}. In \cite{yoo2008asymptotic}, a linear MMSE frequency domain equalizer (FDE) was derived assuming knowledge of the second-order noise statistics. Other work has been focused on cyclostationary noise cancellation at the receiver. In \cite{garcia2007mitigation} and \cite{llano2011quasi}, it was demonstrated the use of linear adaptive predicton filters for cyclostationary noise estimation in NB PLC systems. More recently in \cite{lin2012cyclo}, it was suggested to parameterize the LPTV system model in Fig.\ \ref{fig:cycloTrace} by a periodically switching autoregressive (AR) process, whose switching states and AR parameters can be inferred during training by Bayesian learning. Based on the estimated AR models, a periodically switching moving average (MA) filter can be prepended to the receiver for noise pre-whitening. These filter based algorithms, however, may suffer from the over-fitting problem especially in the simultaneous presence of the asynchronous impulsive noise. To improve the robustness against such outliers generally requires longer training sequences.

Typical cyclostationary noise in PLC systems consists of high power noise bursts that sweep up to $30\%$ of the period \cite{nassar2012cyclostationary}, which is much longer than the duration of an OFDM symbol as specified in the standards.  For example, the OFDM symbol duration in G3 is $231.7\mu s$ in the US Federal Communications Commission (FCC) band, whereas $10\%$ of the noise period (i.e. half the AC cycle) is $833.3\mu s$ in the US. In this situation, noise mitigation algroithms that span multiple OFDM symbols (e.g. a packet) are desirable. If the packet duration is much longer than the noise bursts, there is hope that we can recover the contaminated OFDM symbols using the information from the other impulsive noise free symbols. A time-domain block interleaved OFDM (TDI-OFDM) transceiver structure was proposed in \cite{dweik} to spread long noise bursts into shorter ones over multiple OFDM symbols. After deinterleaving at the receiver, each OFDM symbol is contaminated by a sparse noise vector in the time domain, similarly to the situation under asynchronous impulsive noise. Therefore the non-parametric techniques developed for asynchronous impulsive noise mitigation can also be applied to mitigate cyclostationary noise.

\section{System Model}
\label{sec:SysModel}
\begin{figure}[t]
  \centering
  \centerline{\includegraphics[width=5in]{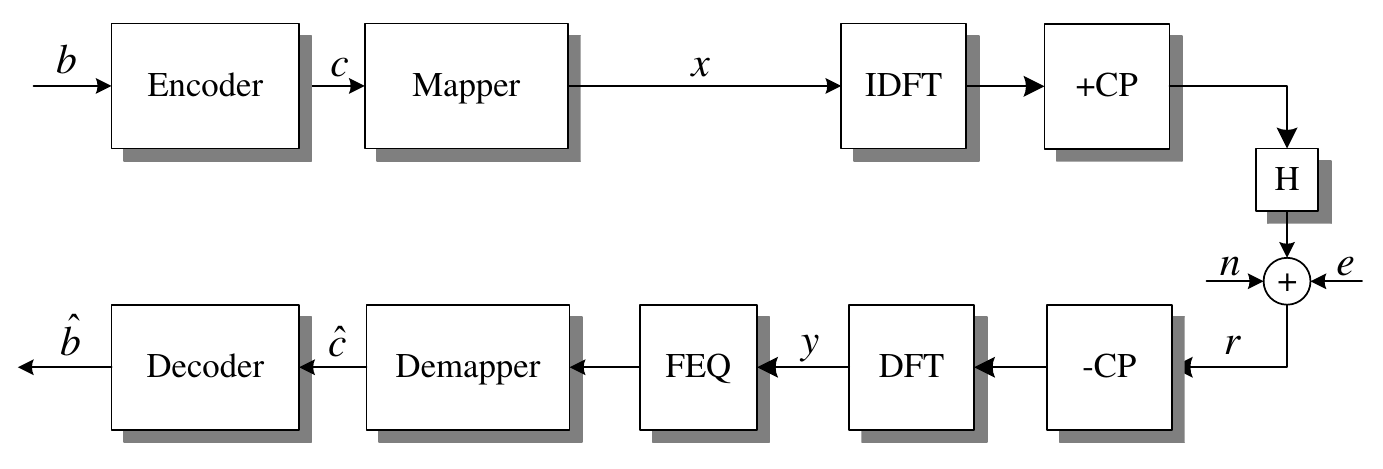}}
\caption{A conventional baseband coded OFDM system.}
\label{fig:ofdmTxRx}
\end{figure}

We first consider a conventional coded OFDM system whose complex baseband equivalent representation is shown in Fig.\ \ref{fig:ofdmTxRx}. At the transmitter, a binary data packet $\mathbf{b}$ is encoded into a codeword $\mathbf{c}$. The codeword is then mapped to OFDM symbols, each with $N-M$ data subcarriers and $M$ non-data subcarriers. The non-data subcarriers are either null tones for spectral shaping and inter-carrier interference reduction, or pilots for channel estimation and synchronization. An OFDM symbol, denoted by $\mathbf{x}$, is converted to the time domain by the inverse DFT (IDFT). A cyclic prefix (CP), assumed to be longer than the channel delay spread, is inserted to the beginning of each OFDM symbol to prevent inter-symbol interference (ISI).

At the receiver, we remove the CP from the received OFDM symbols, resulting in
\begin{equation}
\label{eq:rxSignal}
\mathbf{r}=\mathbf{HF}^*\mathbf{x}+\mathbf{e}+\mathbf{n},
\end{equation}
where $\mathbf{F}$ is the $N$-point DFT matrix, $\mathbf{H}\in\mathbb{C}^{N\times N}$ is the convolutional matrix of the channel and is circulant due to the cyclic prefix insertion, and $\mathbf{e}, \mathbf{n}\in\mathbb{C}^N$ represent impulsive noise and AWGN, respectively. The OFDM demodulator takes the DFT of $\mathbf{r}$, leading to
\begin{eqnarray}\label{eq:demodulatedSignal}
\mathbf{y}&=&\mathbf{FHF}^*\mathbf{x}+ \mathbf{Fe}+\mathbf{Fn} \nonumber \\ 
&=&\mathbf{\Lambda x} + \mathbf{Fe}+\mathbf{g}.
\end{eqnarray}
Here $\mathbf{\Lambda}\triangleq\mathbf{FHF}^*$ is a diagonal matrix, with $\left\{H_i\right\}_{i=1}^N$ (the $N$-point DFT coefficients of the channel impulse response) on its diagonal, and $\mathbf{g}\triangleq\mathbf{Fn}$ is the DFT of $\mathbf{n}$ and is also AWGN since $\mathbf{F}$ is unitary.  

\begin{figure}[t]
  \centering
  \centerline{\includegraphics[width=6in]{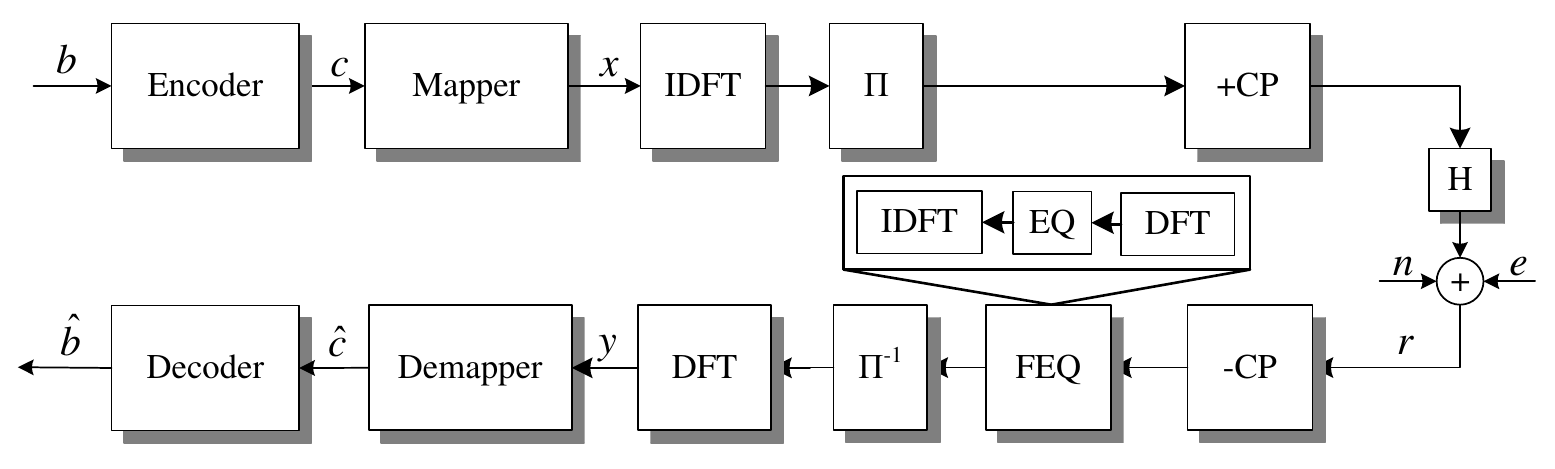}}
\caption{A time-domain interleaved OFDM system. The interleaver $\Pi$ interleaves the OFDM symbols at the sample level while the de-interleaver $\Pi^{-1}$ reverses the operation. The bursty impulsive noise only passes through $\Pi^{-1}$ which spreads the pulses in the time domain and makes them appear sparse to the receiver. }
\label{fig:tdiOfdm}
\end{figure}

In the presence of bursty noise spannning multiple OFDM symbols, we adopt the time-domain block interleaved OFDM transceiver structure as in \cite{dweik} (Fig.\ \ref{fig:tdiOfdm}). At the transmitter we interleave multiple OFDM symbols in the time domain at the sample level before CP insertion. At the receiver, after CP removal, we apply a frequency-domain channel equalizer (FEQ) to the received signal. The equalized signal is then deinterleaved in the time domain before converted to the frequency domain by DFT. Assuming perfect channel estimation, the demodulated OFDM signal $\mathbf{y}$ can be expressed as
\begin{equation}
\mathbf{y'=x+Fe'+g'}.
\label{eq:demodulatedSignal2}
\end{equation}
Here $\mathbf{e'}$ denotes the impulsive noise after deinterleaving, and $\mathbf{g'}$ is spectrally-shaped Gaussian noise due to the FEQ. We suppose that the interleaver size is large enough to spread the long noise bursts into short pulses over multiple OFDM symbols, so that $\mathbf{e'}$ is sparse. For simplicity, we also approximate $\mathbf{g'}$ as AWGN. With these assumptions, \eqref{eq:demodulatedSignal2} can be considered as a special case of \eqref{eq:demodulatedSignal} with a flat channel (i.e. $\boldsymbol{\Lambda}=\mathbf{I}$). Therefore, the following derivations will apply to both \eqref{eq:demodulatedSignal} and \eqref{eq:demodulatedSignal2}.

Let $\mathcal{I}$ denote the index set of the null and pilot tones, where $|\mathcal{I}|= M<N$. Also, let $\left(\cdot \right)_{\mathcal{I}}$ denote the sub-matrix (or sub-vector) corresponding to the rows (or elements) indexed by the set $\mathcal{I}$.  Assuming perfect channel estimation, i.e. complete knowledge of $\Lambda$, the impulsive noise can be observed from the null and pilot tones of the received OFDM symbol, since
\begin{eqnarray}
\mathbf{z}&\triangleq&\mathbf{y_{\mathcal{I}}-(\Lambda x)_{\mathcal{I}}} \nonumber\\ 
                &=&\mathbf{F}_{\mathcal{I}}\mathbf{e}+\mathbf{g}_{\mathcal{I}}, \nonumber \\ 
		\mathbf{g}_{\mathcal{I}}&\sim& \mathcal{CN}(\mathbf{0},\sigma^2\mathbf{I}_M). 
\label{eq:nullTone}
\end{eqnarray}

The recovery of the length-$N$ vector $\mathbf{e}$ from the noisy underdetermined $M\times N$ linear system is generally an ill-conditioned problem. However, exploiting the sparse nature of $\mathbf{e}$ (since impulsive noise has very few non-zero samples in the time domain), we can possibly get an accurate estimate of $\mathbf{e}$ by applying various compressed sensing techniques. 

We would like to use the estimated impulsive noise to improve the detection of $\mathbf{x}$. More specifically, the impulsive noise estimate $\mathbf{\hat{e}}$ can be subtracted out from the received symbol on the data tones to form a new decision metric
\begin{eqnarray}
\hat{\mathbf{y}}_{\overline{\mathcal{I}}} &=& \mathbf{y}_{\overline{\mathcal{I}}} - \mathbf{F}_{\overline{\mathcal{I}}}\mathbf{\hat{e}} \nonumber \\ 
&=& \mathbf{(\Lambda x)}_{\overline{\mathcal{I}}} + \mathbf{g}_{\overline{\mathcal{I}}} + \mathbf{F}_{\overline{\mathcal{I}}}\mathbf{(e-\hat{e})}.
\label{eq:newMetric}
\end{eqnarray}
where $\overline{\left(\cdot\right)}$ indicates set complement and thus $\overline{\mathcal{I}}$ indicates the set of data tone indices.
Assuming that $\mathbf{\hat{e} \approx e}$, the receiver can then proceed as if only Gaussian noise were present and apply the conventional detection and decoding algorithms.

\section{Non-parametric Impulsive Noise Estimation}
\label{sec:impulsiveNoiseEst}
The estimation of impulsive noise converts to solving an underdetermined linear regression problem in \eqref{eq:nullTone} under sparsity constraints. Among various compressed sensing algorithms, sparse Bayesian learning (SBL) has recently obtained much attention since they generally achieve the best recovery performance. Therefore we apply the SBL techniques and propose three non-parametric algorithms for impulsive noise estimation. Before introducing the algorithms, we briefly describe the SBL framework.

\subsection{Sparse Bayesian Learning}
\label{subsec:SBL}
SBL was first proposed by Tipping \cite{tipping2001sparse}, and was introduced to sparse signal recovery by Wipf and Rao in \cite{wipf2004sparse}. Generally, SBL is a Bayesian learning approach for solving the linear regression problem
\begin{equation}
\label{eq:linearRegression}
 \mathbf{t = \Phi w + v},\ \ \mathbf{v}\sim\mathcal{CN}(0,\sigma^2\mathbf{I}_M),
\end{equation} 
where $\mathbf{t}\in\mathcal{C}^M$ is an observation vector, 
$\mathbf{\Phi}= \begin{bmatrix} 
\Phi_1 & \cdots & \Phi_N
\end{bmatrix}
\in \mathcal{C}^{M\times N}$ 
is an overcomplete basis (i.e. $M<N$), and $\mathbf{w}\in \mathcal{C}^N$ is a sparse weight vector to be estimated. 

SBL imposes a parameterized Gaussian prior on $\mathbf{w}$ 
\begin{equation}
\label{eq:SBLprior}
p(\mathbf{w} ; \boldsymbol{\Gamma})=\mathcal{CN}(\mathbf{w} ; \mathbf{0}, \boldsymbol{\Gamma}),
\end{equation}
where $\boldsymbol{\Gamma}\triangleq diag\{\boldsymbol{\gamma}\}$, and $\boldsymbol{\gamma}\in\mathcal{R}^N$ whose $i$-th component $\gamma_i$ is the variance of $w_i$. Given the prior, the likelihood of the observation can be expressed as
\begin{equation}
\label{eq:LL}
p(\mathbf{t}; \boldsymbol{\Gamma},\sigma^2)=\mathcal{CN}(\mathbf{t}; \mathbf{0}, \boldsymbol{\Phi\Gamma\Phi^*}+\sigma^2\mathbf{I}_M).
\end{equation}
A maximum likelihood (ML) estimator solves the hyperparameters $\boldsymbol{\gamma}$ and $\mathbf{\sigma}^2$ that maximize \eqref{eq:LL}. The ML estimation is computed iteratively using expectation maximization (EM).

Given the observations and the estimated hyperparameters, the posterior density of $\mathbf{e}$ is also a Gaussian distribution
\begin{eqnarray}
\label{eq:posterior}
p(\mathbf{w}|\mathbf{t};\boldsymbol{\Gamma},\sigma^2)&=&\mathcal{CN}(\mathbf{w}; \boldsymbol{\mu_w},\boldsymbol{\Sigma_w}), \nonumber \\ 
\boldsymbol{\mu_w}&=&\sigma^{-2}\boldsymbol{\Sigma_w\Phi}^*\mathbf{t}, \nonumber\\ 
\boldsymbol{\Sigma_w}&=&(\sigma^{-2}\boldsymbol{\Phi^*\Phi+\Gamma^{-1}})^{-1}.
\end{eqnarray}
The maximum \textit{a posteriori} (MAP) estimate of $\mathbf{w}$ is the posterior mean $\boldsymbol{\mu_w}$.

Due to the sparsity promoting property of the prior, upon convergence most components of $\boldsymbol{\gamma}$ and hence $\boldsymbol{\mu_w}$ are driven to zero, rendering a sparse estimate of $\mathbf{w}$. SBL has been proven to be more robust compared to the Basis Pursuit \cite{chen2001atomic} and FOCUSS \cite{gorodnitsky2002sparse} algorithms, since the global optimum is always the sparsest solution, all local optimal solutions are sparse, and the number of local optima is the smallest.

\subsection{Estimation Using Null and Pilot Tones}
\label{subsec:Est1}
The SBL technique can be directly applied to the impulsive noise estimation using null and pilot tones, since substituting $\mathbf{t=z}$, $\boldsymbol{\Phi}=\mathbf{F_\mathcal{I}}$, $\mathbf{w=e}$, and $\mathbf{v=g_\mathcal{I}}$ into \eqref{eq:linearRegression} gives exactly \eqref{eq:nullTone}.

To obtain the ML estimates of the hyperparameters $\boldsymbol{\gamma}$ and $\sigma^2$, we treat $\mathbf{e}$ as the latent variable and apply the EM algorithm. The update of the hyperparameters in the $k$-th iteration is as follows, where we define $\boldsymbol{\theta}\triangleq(\boldsymbol{\gamma},\sigma^2)$ for conciseness.
\begin{eqnarray}
	\gamma_i^{(k+1)}&=&\underset{\gamma_i\geq 0}{\text{argmax}}\ \mathbb{E}_{\mathbf{e}|\mathbf{z};\boldsymbol{\theta}^{(k)}}[\log\ p(\mathbf{z},\mathbf{e};\boldsymbol{\theta}^{(k)})]\notag\\
	&=&\mathbb{E}_{\mathbf{e}|\mathbf{z};\boldsymbol{\theta}^{(k)}}[\mathbf{e}_i^2]\notag\\
	&=&\mathbf{\Sigma_e}_{,ii}^{(k)}+(\boldsymbol{\mu_e}_{,i}^{(k)})^2,
\label{eq:gamma}\\
	(\sigma^2)^{(k+1)}&=&\frac{1}{M}\{||\mathbf{z}-\mathbf{F}_{\mathcal{I}}\boldsymbol{\mu_e}^{(k)}||^2 +\nonumber\\
	&&(\sigma^2)^{(k)}\sum_{i=1}^N[1-(\gamma_i^{(k)})^{-1}\mathbf{\Sigma_{e}}^{(k)}_{,ii}]\},
\label{eq:sigma}\\
       \boldsymbol{\mu_e}^{(k)}&=&(\sigma^{-2})^{(k)}\boldsymbol{\Sigma_e}^{(k)} \mathbf{F}_\mathcal{I}^*\mathbf{z}, 
\label{eq:mu}\\
        \boldsymbol{\Sigma_e}^{(k+1)}&=&[ (\sigma^{-2})^{(k)}\mathbf{F}_\mathcal{I}^*\mathbf{F}_\mathcal{I}+(\boldsymbol{\Gamma}^{(k)})^{-1}]^{-1}.
\label{eq:Sigma}
\end{eqnarray}

Upon termination of the EM algorithm, we obtain the MAP estimate of the time-domain impulsive noise $\hat{\mathbf{e}}=\boldsymbol{\mu_e}$. We then transform $\hat{\mathbf{e}}$ to the frequency domain and subtract it from the received signal on the data tones according to \eqref{eq:newMetric}. 

\subsection{Estimation Using All Tones}
\label{subsec:Est2}
As will be demonstrated in the simulation results, performance of the estimator using null and pilot tones is improved as the number of these non-data tones increases. However having fewer data tones means reduced throughput. When the number of non-data tones is limited, it is desirable to exploit information available in all tones to estimate the impulsive noise. To do this, we define $\mathbf{u}\triangleq \boldsymbol{\Lambda}\mathbf{x+g}$, and rewrite \eqref{eq:demodulatedSignal} as

\begin{eqnarray}
\left[ {\begin{array}{c}
 \mathbf{z}  \\
 \mathbf{y}_{\overline{\mathcal{I}}}  \\
 \end{array} } \right]
&=&\mathbf{Fe}+
\left[ {\begin{array}{c}
 \mathbf{u}_{\mathcal{I}}  \nonumber \\
 \mathbf{u}_{\overline{\mathcal{I}}}  \\
 \end{array} } \right], \nonumber \\ 
\mathbf{u}_{\mathcal{I}}&\sim& \mathcal{CN}(\mathbf{0},\sigma^2 \mathbf{I}_M),\nonumber \\ 
\mathbf{u}_{\overline{\mathcal{I}}}&\sim& \mathcal{CN}((\boldsymbol{\Lambda x})_{\overline{\mathcal{I}}},\sigma^2 \mathbf{I}_{N-M}).
\end{eqnarray}

Imposing the parameterized Gaussian prior on $\mathbf{e}$, i.e. $p(\mathbf{e};\boldsymbol{\Gamma})=\mathcal{CN}(\mathbf{0},\boldsymbol{\Gamma})$, the likelihood of $\mathbf{z}$ remains the same as \eqref{eq:LL}, while the likelihood of $\mathbf{y}_{\overline{\mathcal{I}}}$, the received signal on the data tones, is
\begin{eqnarray}
\label{eq:LL2}
p\big(\mathbf{y}_{\overline{\mathcal{I}}}; (\boldsymbol{\Lambda x})_{\overline{\mathcal{I}}},\boldsymbol{\Gamma},\sigma^2\big)&=&\mathcal{CN}\big(\mathbf{y}_{\overline{\mathcal{I}}}; (\boldsymbol{\Lambda x})_{\overline{\mathcal{I}}}, \boldsymbol{\Sigma}_{\mathbf{y}_{\overline{\mathcal{I}}}}\big), \nonumber \\ 
\boldsymbol{\Sigma}_{\mathbf{y}_{\overline{\mathcal{I}}}}&=&\boldsymbol{F}_{\overline{\mathcal{I}}}\boldsymbol{\Gamma F}_{\overline{\mathcal{I}}}^*+\sigma^2\mathbf{I}_{N-M},
\end{eqnarray}
with the unknown transmitted signal $\mathbf{x}_{\overline{\mathcal{I}}}$ as a third hyperparameter in addition to $\boldsymbol{\Gamma}$ and $\sigma^2$. Although $\mathbf{x}_{\overline{\mathcal{I}}}$ consists of constellation points, which are discrete, we temporarily relax it to be continuous when treated as a hyperparameter in the EM algorithm. 

The iterative updates in the EM algorithm now involve all three hyperparameters. Since each hyperparameter is updated while keeping the others fixed, the update equations for $\boldsymbol{\gamma}$ and $\sigma^2$ are in the same forms as \eqref{eq:gamma} and \eqref{eq:sigma}. We treat ($\boldsymbol{\Lambda x})_{\overline{\mathcal{I}}}$ in a whole as a hyperparameter and update it as
\begin{eqnarray}
(\boldsymbol{\Lambda x})_{\overline{\mathcal{I}}}^{(k+1)}&=&\underset{(\boldsymbol{\Lambda x})_{\overline{\mathcal{I}}}}{\text{argmax}}\ \mathbb{E}_{\mathbf{e}|\mathbf{y};\boldsymbol{\theta}^{(k)}}[\log\ p(\mathbf{y,e};\boldsymbol{\theta}^{(k)})]\notag\\
&=&\underset{(\boldsymbol{\Lambda x})_{\overline{\mathcal{I}}}}{\text{argmin}}\ |\mathbf{y}_{\overline{\mathcal{I}}}-(\boldsymbol{\Lambda x})_{\overline{\mathcal{I}}}-\mathbf{F}_{\overline{\mathcal{I}}}\boldsymbol{\mu_e}^{(k)}|^2 \notag\\
&=&\mathbf{y}_{\overline{\mathcal{I}}}-\mathbf{F}_{\overline{\mathcal{I}}}\boldsymbol{\mu_e}^{(k)}.
\end{eqnarray}
The entire EM algorithm is summarized as follows.
\begin{eqnarray}
	\gamma_i^{(k+1)}&=&\underset{\gamma_i\geq 0}{\text{argmax}}\ \mathbb{E}_{\mathbf{e}|\mathbf{y};\boldsymbol{\theta}^{(k)}}[\log\ p(\mathbf{y,e};\boldsymbol{\theta}^{(k)})]\notag\\
	&=&\mathbf{\Sigma_e}_{,ii}^{(k)}+(\boldsymbol{\mu_e}_{,i}^{(k)})^2,
\label{eq:gammaJoint}\\
	(\sigma^2)^{(k+1)}&=&\frac{1}{N}(||\mathbf{y-\Lambda x}^{(k)}-\mathbf{F}\boldsymbol{\mu}^{(k)}||^2
				 +\nonumber\\
				&&(\sigma^2)^{(k)}\sum_{i=1}^N[1-(\gamma_i^{(k)})^{-1}\mathbf{\Sigma_e}_{,ii}^{(k)},
\label{eq:sigmaJoint}\\
	(\mathbf{\Lambda x})_{\overline{\mathcal{I}}}^{(k+1)}&=&\mathbf{y_{\overline{\mathcal{I}}}-F_{\overline{\mathcal{I}}}e}^{(k)},
\label{eq:xJoint}\\
\mathbf{\Sigma_e}^{(k)}&=&\mathbf{\Gamma}^{(k)}-\mathbf{\Gamma}^{(k)}\mathbf{F}^*\Sigma_{\mathbf{y}}^{-1}\mathbf{F}\Gamma^{(k)},
\label{eq:sigma_eJoint}\\
\hat{\mathbf{e}}^{(k)}&=&\boldsymbol{\mu_e}^{(k)}= \frac{1}{\left(\sigma^2\right)^{(k)}} \mathbf{\Sigma_e}^{(k)}\mathbf{F}^*(\mathbf{y}-\mathbf{\Lambda x}^{(k)}).
\label{eq:muJoint}
\end{eqnarray}

\subsection{Decision Feedback Estimation}
\label{subsec:DFE}

The two estimators described above impose a parameterized Gaussian prior on the time-domain impulsive noise, i.e. $\mathbf{e}\sim \mathcal{CN}(0,\boldsymbol{\Gamma})$. Prior information on $\boldsymbol{\Gamma}$, or equivalently on the precision matrix $\mathbf{P}\triangleq\boldsymbol{\Gamma}^{-1}$, can be introduced by the conjugate prior distribution on these hyperparameters. Let $\mathbf{p}\triangleq[p_1,\cdots,p_N]^T$ denote the diagonal of $\mathbf{P}$. The conjugate prior on $\mathbf{p}$ is a Gamma distribution 
\begin{equation}
\label{eq:gammaPrior}
P(\mathbf{p;a,b})=\prod\limits_{i=1}^N \text{Ga}(p_i;a_i,b_i).
\end{equation}
where $\text{Ga}\left(\cdot ;a,b \right)$ is the Gamma distribution with parameters $a$ and $b$.
When $a_i=0, b_i=0, \forall i$, \eqref{eq:gammaPrior} reduces to a uniform distribution, which is an uninformative prior that is implicitly imposed in the previously described SBL framework. Non-zero values of $a_i$ and $b_i$ contain prior information that is integrated into the likelihood function in \eqref{eq:LL}, resulting in
\begin{eqnarray}
\label{eq:LL3}
p(\mathbf{z}; \mathbf{P},\sigma^2,\mathbf{a},\mathbf{b})&=&\mathcal{CN}(\mathbf{z};\mathbf{0}, \mathbf{F}_{\mathcal{I}}\mathbf{P}^{-1}\mathbf{F}_{\mathcal{I}}^*+\sigma^2\mathbf{I}_M)\cdot\nonumber \\
&& \text{Ga}(\mathbf{P};\mathbf{a,b}).
\end{eqnarray}
Differentiating \eqref{eq:LL3} with respect to $\log p_i$, and setting the derivatives to zeros, we obtain
\begin{equation}
\label{eq:updateGamma}
p_i = \gamma_i^{-1} = \frac{1+2a_i}{\boldsymbol{\mu_e}_{,i}^2+\boldsymbol{\Sigma_e}_{,ii}+2b_i}.
\end{equation}
Comparing \eqref{eq:updateGamma} to \eqref{eq:gamma}, we can see the prior information contained in $a_i$ and $b_i$ does affect the ML estimates of $\boldsymbol{\gamma}$.
Since \eqref{eq:updateGamma} is the conjugate prior on $\mathbf{p}$, the posterior probability of $\mathbf{p}$ given $\mathbf{e, a}$ and $\mathbf{b}$ is also Gamma distributed, i.e.
\begin{equation}
\label{eq:gammaPost}
P(\mathbf{p}|\mathbf{e;a,b}) = \prod_{i=1}^N\text{Ga}(p_i;\tilde{a}_i,\tilde{b}_i)
\end{equation}
with the updated parameters
\begin{eqnarray}
\label{eq:update}
\tilde{a}_i &=& a_i+\frac{1}{2}, \notag\\ 
\tilde{b}_i &=& b_i+\frac{|e_i|^2}{2}.
\end{eqnarray}

Suppose that in addition to the MAP estimate $\hat{\mathbf{e}}$ given by the estimator using non-data tones, a second estimate of $\mathbf{e}$, denoted by $\tilde{\mathbf{e}}$, is available based on certain side information. The side information contained in $\tilde{\mathbf{e}}$ can be fused into $\hat{\mathbf{e}}$ via the posterior distribution of $\mathbf{p}$ given $\tilde{\mathbf{e}}$. More specifically, given $\tilde{\mathbf{e}}$, we update $\mathbf{a}$ and $\mathbf{b}$ according to \eqref{eq:update}, and then solve the ML estimate of $\mathbf{p}$ \eqref{eq:updateGamma} with the updated $\tilde{\mathbf{a}}$ and $\tilde{\mathbf{b}}$.

\begin{figure}[t]
  \centering
  \includegraphics[width=5.5in]{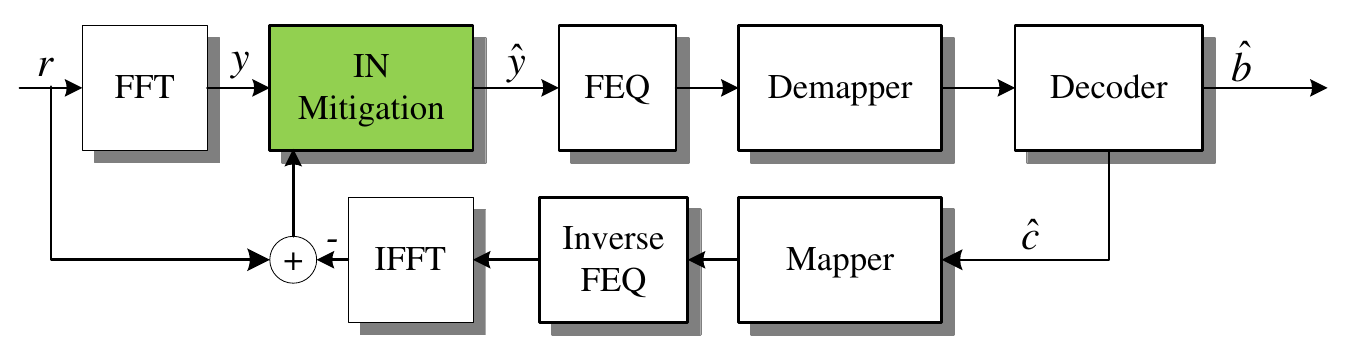}
\caption{A decision feedback impulsive noise estimator.}
\label{fig:turboSBL}
\end{figure}

In coded OFDM systems, the redundancy in the coded data tones can be exploited as the side information to provide a second estimate of $\tilde{\mathbf{e}}$. More specifically, the decoder takes the OFDM symbols after impulsive noise mitigation as the input, and produces hard decisions on the uncoded and coded bits, $\hat{\mathbf{b}}$ and $\hat{\mathbf{c}}$, respectively. Using $\hat{\mathbf{c}}$ we can recover the data tones of the OFDM symbols by appropriate constellation mapping. This gives an estimate of $\hat{\mathbf{x}}_{\mathcal{\overline{I}}}$, which is multiplied by the channel frequency response $\boldsymbol{\Lambda}$, transformed to the time domain and subtracted from the received signal $\mathbf{r}$ to generate the estimate $\tilde{\mathbf{e}}$. Then we use $\tilde{\mathbf{e}}$ to update $\mathbf{a}$ and $\mathbf{b}$, through which the information extracted from the coding redundancy is transferred back to the impulsive noise estimator. As such, we form a decision feedback estimator that transfers information back-and-forth between the impulsive noise estimator using non-data tones and the decoder using data tones (Fig.\ \ref{fig:turboSBL}). Compared to the estimator using all tones in Section \ref{subsec:Est2}, the decision feedback estimator is expected to have better performance by exploiting the redundant information (i.e. coding structure) on the data tones.

\section{Low-Complexity Implementation}
\label{sec:SequentialSBLImplementation}
The core SBL algorithm in \eqref{eq:posterior} involves an $M\times M$ matrix inversion ($M$ is the number of null and pilot tones) which might be practically infeasible on a hardware platform. An accelerated  version of SBL that utilizes the properties of the marginal likelihood has been proposed in \cite{Tipping2003}. Using these properties, the accelerated SBL algorithm performs a sequential addition and deletion of candidate basis functions given by the columns of $\mathbf{\Phi}$ in \eqref{eq:linearRegression}, while keeping the same reconstruction performance \cite{Tipping2003}.
\begin{algorithm}[h]
\caption{Sequential SBL algorithm \cite{Tipping2003}}\label{alg:seqSBL}
\begin{algorithmic}[1]
\State Initialize background noise variance $\sigma^2$
\State Select $\mathbf{f}_0$ with largest projection $\left\|\mathbf{f}_0 \mathbf{y}\right\|^2$ on the observed vector $\mathbf{y}$
\State Compute $\alpha_0 = \left(\left\|\mathbf{f}_0 \mathbf{y}\right\|^2 - \sigma^2 \right)^{-1}$, all other $\alpha_j=\infty$ (exclude from model)
\State Compute $\boldsymbol{\Sigma}$ and $\boldsymbol{\mu}$ as given by \cite{Tipping2003}
\While{$\Delta \alpha_j \leq$ threshold , $\forall j$}
\State Select a candidate basis $\mathbf{f}_i$ from columns of $F$
\State Compute $\theta_i=\left\|q_i\right\|^2- s_i$
\If {$\theta_i>0$ and $\alpha_i<\infty$}
\State $\mathbf{f}_i$ is in model, re-estimate $\alpha_i$ as given in \cite{Tipping2003} \label{alg:restimate}
\ElsIf{$\theta_i>0$ and $\alpha_i=\infty$}
\State $\mathbf{f}_i$ not in model, add $\mathbf{f}_i$ and update parameters \label{alg:add}
\Else
\State $\mathbf{f}_i$ is in model, remove $\mathbf{f}_i$ and update parameters \label{alg:remove}
\EndIf
\State Update noise variance $\sigma^2$ as given in \cite{Tipping2003}
\EndWhile
\end{algorithmic}
\end{algorithm}

\begin{table}[b]
  \centering
  	\begin{tabular}{ | c | c | c | }
  	\hline
  	\bfseries{Estimator} & \bfseries{Operation} & \bfseries{Complexity} \\ \hline
  	\multirow{2}{*}{Using null and pilot Tones} & Matrix multiply & $\mathcal{O}(N^2M)$ \\ \cline{2-3}
	& Matrix inversion & $\mathcal{O}(M^3)$ \\ \hline
  	\multirow{2}{*}{Using all tones} & Matrix multiply & $\mathcal{O}(N^3)$ \\ \cline{2-3}
	& Matrix inversion & $\mathcal{O}(N^3)$ \\ \hline
  	Sequential SBL  & Matrix multiply  & $\mathcal{O}(N^2K)$  \\ \cline{2-3}
	w/ unknown background noise power &  Matrix inversion & $\mathcal{O}(K^3)$ \\ \hline
	Sequential SBL  & \multirow{2}{*}{Matrix multiply}  & \multirow{2}{*}{$\mathcal{O}(N^2K)$} \\ 
	w/ known background noise power &   & \\ \hline
  	\end{tabular}
  \caption{Complexity per iteration of the proposed algorithms. $N$ is the FFT size, $M$ is the number of null and pilot tones, and $K$ is the number of model basis in the current iteration.}
  \label{tab:complexity}
\end{table}

In the context of impulsive noise estimation, an impulse at noise sample $i$ is represented as a non-zero entry in vector $\mathbf{e}$ at index $i$. As a result, it contributes $e_i \mathbf{f}_i$ to the observed output $\mathbf{y}$, where we denote the $i$-th column in $\mathbf{F}_\mathcal{I}$ as $\mathbf{f}_i$. The  accelerated algorithm will sequentially add, remove, or update basis $\mathbf{f}_i$ until convergence. At convergence, the basis that remain in the model will indicate the support of vector $\mathbf{e}$ and thereby the locations of the impulses. Algorithm \ref{alg:seqSBL} presents a high-level description of the sequential algorithm. Due to the space constraints, we refer interested readers to \cite{Tipping2003} for the mathematical details.

Significant computational savings can be obtained if the background noise power $\sigma^2$ is known.  This will allow for  efficient calculations in steps \ref{alg:restimate}, \ref{alg:add}, \ref{alg:remove} in algorithm \ref{alg:seqSBL} without any matrix inversions as given in \cite{Tipping2003}. \tablename~\ref{tab:complexity} compares the complexity per iteration of the original SBL-based algorithms and the sequential implementation of it.
The computational complexity of the original algorithms is dominated by the matrix multiplication and inversion operations in (\ref{eq:sigma}) and (\ref{eq:sigmaJoint}). 
Compared to the estimator using non-data tones, the estimator using all tones increases the complexity from $\mathcal{O}(N^2M)$ per iteration to $\mathcal{O}(N^3)$ per iteration, where $N$ is the DFT size. On the other hand, each iteration of the sequential SBL involves matrix multiplications and inversions that have complexities of $\mathcal{O}(N^2K)$ and $\mathcal{O}(K^3)$, respectively, where $K$ is the number of model basis in that particular iteration. Furthermore, it can make use of the knowledge of background noise power to eliminate any matrix inversion operations.

\section{Simulation Results}
\label{sec:SimulationResults}
\begin{table}[t]
  \centering
  	\begin{tabular}{ | c | c | c | }
  	\hline
  	\bfseries{Parameters} & \bfseries{Simulation} & \bfseries{G3 Standard} \\ \hline
	FFT Length & 128 & 256 \\ \hline
  	Modulation & QPSK & DPSK ($|\mathcal{C}|$=2,4,8) \\ \hline
	$\#$ of Tones & 128 & 128 \\ \hline
	$\#$ of Data Tones & 72 ($\#33-\#104$) & 72 ($\#33-\#104$) \\ \hline
	$\#$ of Null Tones & 56 & 56 \\ \hline
	Forward Error & \multirow{2}{*}{Rate-1/2 Convolutional} & Rate-1/2 Convolutional \\
	Correction Code &  & + Reed Solomon\\ \hline
	\multirow{2}{*}{Interleaver} & Random interleaver & s-interleaver \\
       & in time domain & In frequency domain \\ \hline
	Interleave Size & 100 OFDM symbols & Entire packet \\ \hline
  	\end{tabular}
  \caption{Parameters of the simulated OFDM system and the G3 standard. $|\mathcal{C}|$ denotes the constellation size.}
  \label{tab:params}
\end{table}

To quantify the performance of our proposed algorithms, we simulate a baseband OFDM system over a flat channel. Some parameters of the OFDM system are listed and compared to the G3 PLC standard in Table \ref{tab:params}.

We generate the asynchronous impulsive noise from two different statistical models: a 3-component Gaussian mixture (GM) distribution with $\boldsymbol{\pi}=[0.9,0.07,0.03]$ and $\boldsymbol{\gamma}=[1,100,1000]$, and a Middleton Class A (MCA) distribution with $A=0.1$, $\Omega=0.01$, and the probability density function truncated to the first 10 mixture components. The noise samples are assumed to be i.i.d. In addition, we synthesize cyclostationary noise using the LPTV model proposed in \cite{nassar2012cyclostationary}, with the filter spectrums as in Fig.\ \ref{fig:cycloTrace}.

\begin{figure*}[t]
\begin{center}$
\begin{array}{cc}
\includegraphics[width=3in]{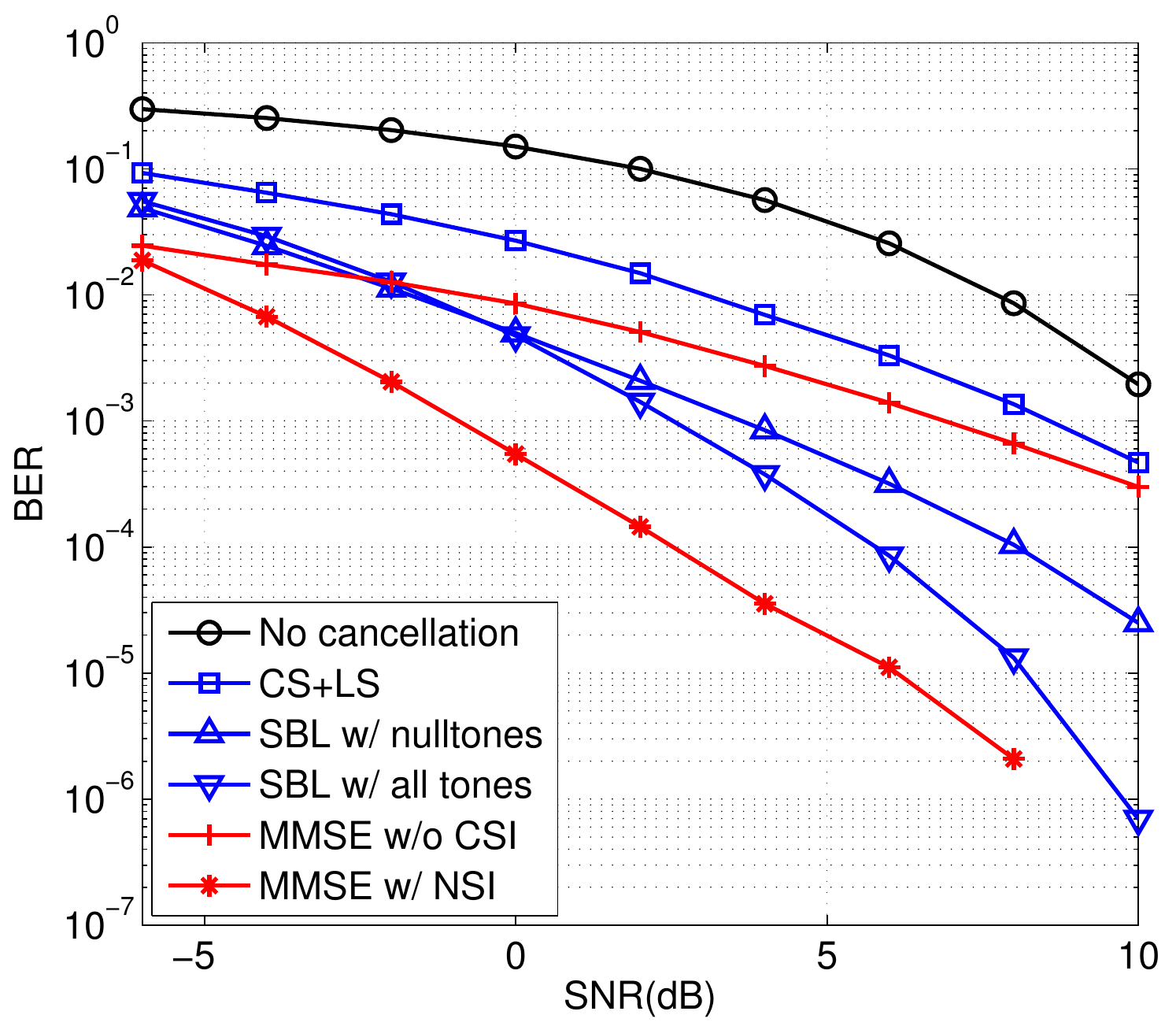} &
\includegraphics[width=3in]{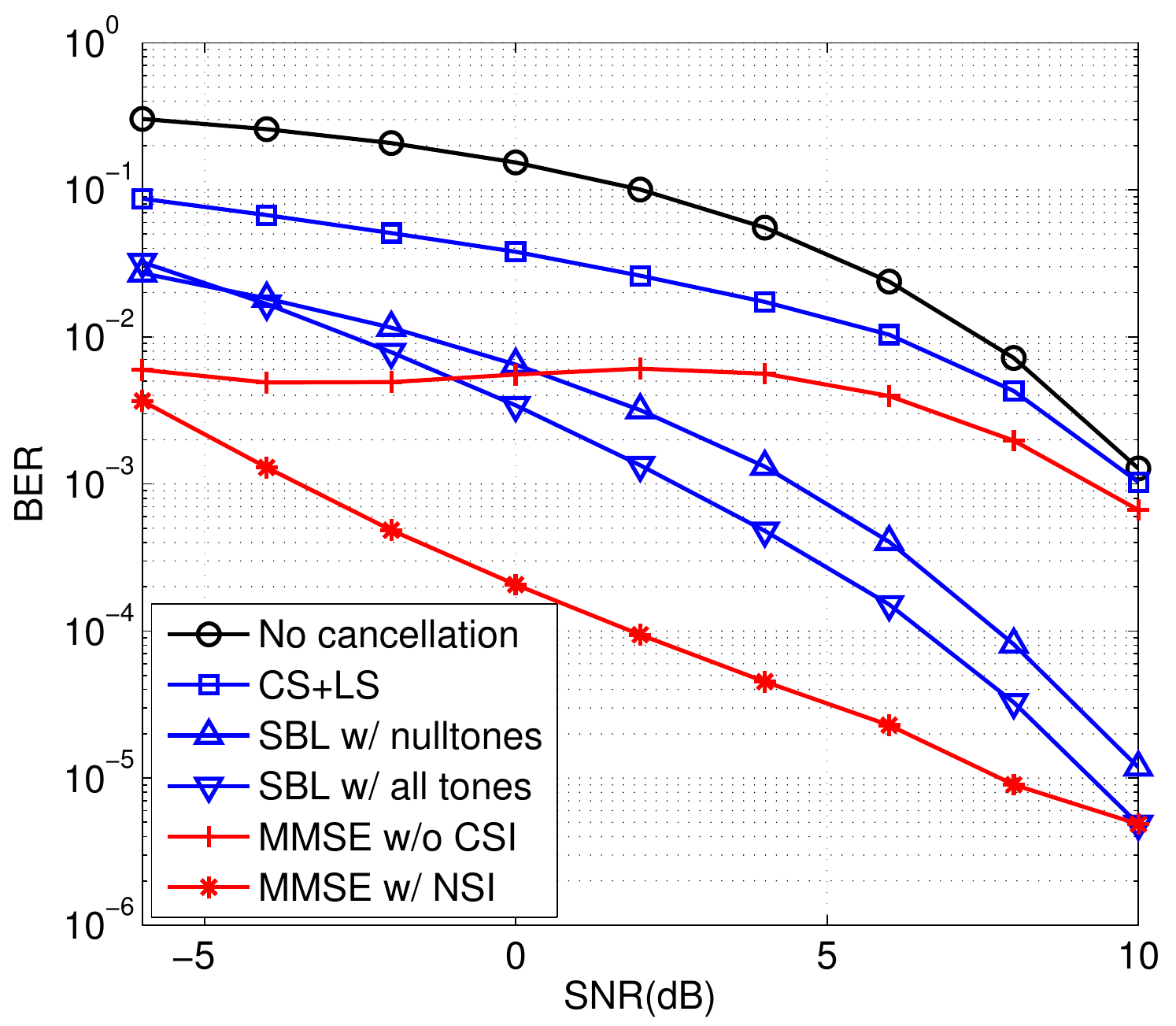}\\
\includegraphics[width=3in]{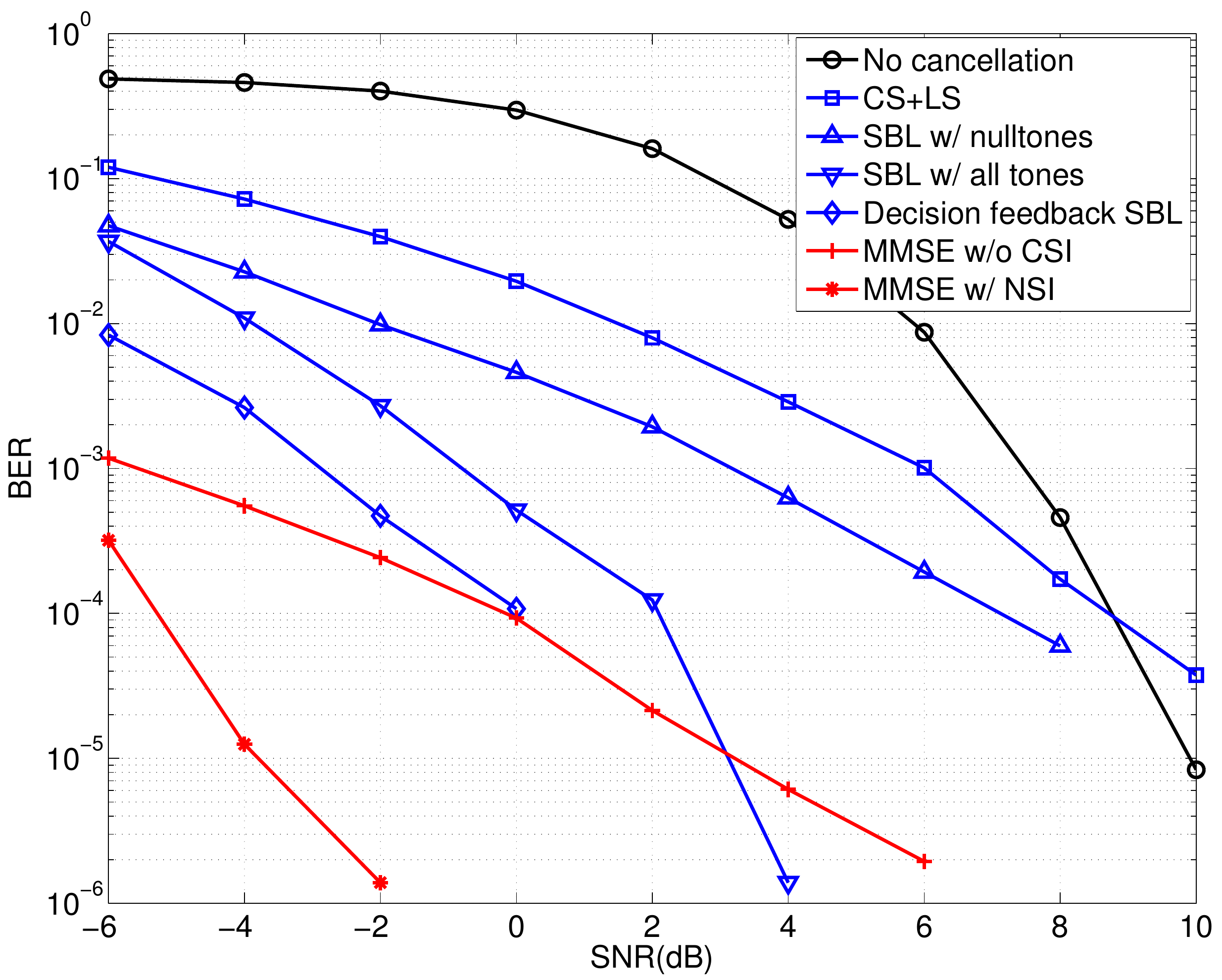} &
\includegraphics[width=3in]{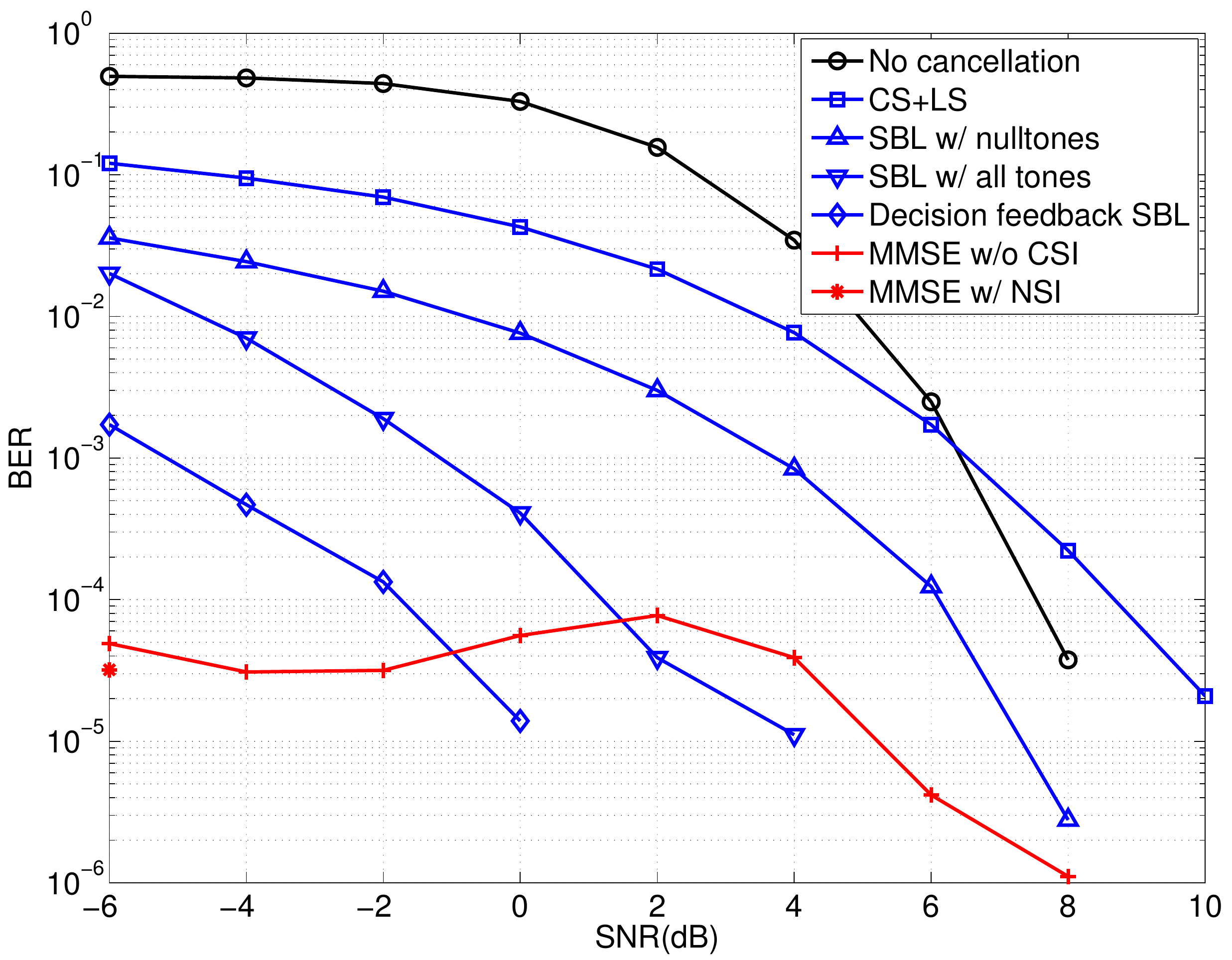}

\end{array}$
\end{center}
\caption{Uncoded (top row) and coded (bottom row) BER performance of proposed algorithms in Gaussian mixture (left column) and Middleton Class A (right column) modeled asynchronous impulsive noise, in comparison with the conventional OFDM system without denoising, the compressed sensing based algorithm \cite{caire2008impulse}, and two parametric MMSE estimators \cite{haringThesis}.}
\label{fig:BER}
\end{figure*}

\begin{figure}[h]
\begin{center}
\includegraphics[width=4in]{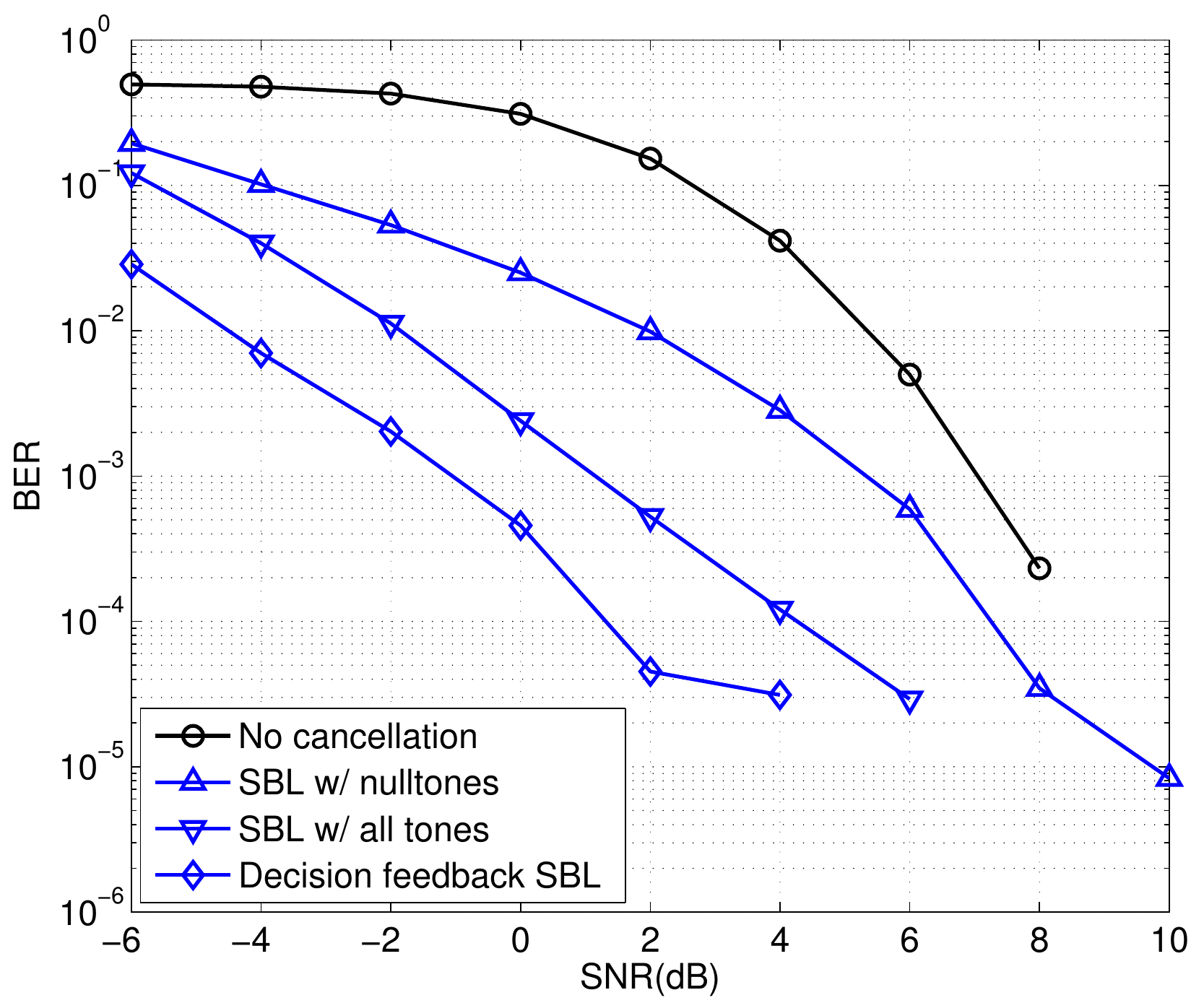} 
\end{center}
\caption{Coded BER performance of proposed algorithms in cyclostationary impulsive noise.}
\label{fig:codedBERCyclic}
\end{figure}

We compare our proposed algorithms with the non-parametric CS and least squares based algorithm in \cite{caire2008impulse} (denoted as CS+LS in the following). The two parametric MMSE detectors in \cite{haringThesis} are also implemented to provide baseline references in the asynchronous impulsive noise. Both MMSE detectors assume perfect knowledge of the Gaussian mixture impulsive noise model, and one even assumes perfect knowledge of the noise variance at each time instance (i.e. noise state information, or NSI). The algorithms are simulated with the convolutional code switched on and off. We measure the bit error rates (BER) over $10^4$ OFDM symbols, which contain $1.44\times 10^6$ and $7.2\times 10^5$ bits in uncoded and coded systems, respectively. The BER performance of the algorithms in uncoded and coded systems in different impulsive noise scenarios are plotted in Fig.\ \ref{fig:BER} and \ref{fig:codedBERCyclic}.

\begin{table}[b]
  \centering
  	\begin{tabular}{ | c | c | c | c | c | }
  	\hline
  	 \multirow{2}{*}{\bfseries{System}} & \multirow{2}{*}{\bfseries{Noise}} & \bfseries{SBL} & \bfseries{SBL} & \bfseries{SBL}\\
	& & \bfseries{w/ Null Tones} & \bfseries{w/ All Tones} & \bfseries{w/ Feedback} \\ \hline
	\multirow{2}{*}{Uncoded} & GM & 8~dB & 10~dB & - \\ \cline{2-5}
  	& MCA & 6~dB & 7~dB &  - \\ \hline
	\multirow{3}{*}{Coded} & GM & 2~dB & 7~dB & 9~dB \\ \cline{2-5}
  	& MCA & 1.75~dB & 6.75~dB & 8.75~dB \\ \cline{2-5}
	& Cyclic & 2~dB & 5~dB & 9.25~dB \\ \hline
  	\end{tabular}
  \caption{SNR gains of the proposed impulsive noise mitigation algorithms over the conventional OFDM system without denoising.}
  \label{tab:SNRgains}
\end{table}

In the uncoded system, our proposed estimator using null tones achieves 6--8~dB SNR gain over conventional OFDM receivers in the asynchronous impulsive noise. We have got additional 1--2~dB gain in a relatively wide SNR region by using all tones. We notice a marginal performance loss of the estimator using all tones in low SNR regimes. This is because at lower SNRs, the errors introduced by the continuous relaxation of constellation points $\mathbf{x}$ (see Section \ref{subsec:Est2}) become more significant. Compared to the parametric MMSE estimators, our proposed estimators outperform the MMSE estimator without NSI in moderate and high SNR regimes.

In the coded system, the proposed estimator using null tones can achieve up to 10 dB SNR gain over conventional OFDM receivers in the asynchronous impulsive noise. The estimator using all tones provides an additional 2--5~dB gains. Furthermore, using decision feedback from the convolutional decoder, we obtain an extra 2~dB gain. In the cyclostationary impulsive noise, we achieve around 5~dB SNR gain with the estimator using null tones, and an additional 2--4~dB gains by using all tones. The decision feedback estimator in this case has an extra 4~dB gain over the one using all tones.

For clarity purposes, we measured the approximate SNR gains of the proposed algorithms over the conventional OFDM system without denoising at a target BER of $10^{-4}$, as listed in TABLE \ref{tab:SNRgains}.

In Fig.\ \ref{fig:BER}, we can see that the compressed sensing based (CS+LS) algorithm performs worse than our proposed estimators in both noise scenarios. As mentioned previously in Section \ref{sec:RelatedWork}, this is because the CS algorithm can only recover the impulsive noise with high sparsity, i.e. typically less than 5 impulses per OFDM symbol in our system settings.

\section{Conclusion}
This paper proposes three methods for improving communication performance of OFDM PLC systems in the presence of asynchronous impulsive noise and cyclostationary impulsive noise. To mitigate asynchronous impulsive noise, we apply sparse Bayesian learning (SBL) techniques to estimate the impulsive noise from the received signal by observing information either on the null and pilot subcarriers or on all subcarriers. Under cyclostationary noise, we adopt a time-domain interleaving OFDM transceiver structure to break long noise bursts that span multiple OFDM symbols into short bursts, and then apply the SBL techniques. All the methods are non-parametric, i.e. do not require prior knowledge on the statistical noise model or model parameters. We validate the proposed algorithms based on Gaussian mixture and Middleton Class A modeled asynchronous impulsive noise, and the cyclostationary impulsive noise.


%


\ifCLASSOPTIONcaptionsoff
  \newpage
\fi



\bibliographystyle{IEEEtranTCOM}
\bibliography{paper}
%
%


%




\end{document}